\documentclass[conference]{IEEEtran}

\IEEEoverridecommandlockouts
\usepackage{cite}
\usepackage{amsmath,amssymb,amsfonts}
\usepackage{algorithmic}
\usepackage{graphicx}
\usepackage{textcomp}
\usepackage{xcolor}
\usepackage{balance}
\def\BibTeX{{\rm B\kern-.05em{\sc i\kern-.025em b}\kern-.08em
    T\kern-.1667em\lower.7ex\hbox{E}\kern-.125emX}}

\usepackage{comment}
    
\begin{document}

\title{``Hello, I'm Delivering. Let Me Pass By'':\\Navigating Public Pathways with Walk-along with Robots in Crowded City Streets 

}


\author{
\IEEEauthorblockN{EunJeong Cheon}
\IEEEauthorblockA{\textit{School of Information Studies}, 
\textit{Syracuse University} 
\\
echeon@syr.edu}
\and

\IEEEauthorblockN{Do Yeon Shin}
\IEEEauthorblockA{\textit{Department of Anthropology}, 
\textit{University of Illinois, Chicago} 
 \\
dshin34@uic.edu}
}
\maketitle

\begin{abstract}
As the presence of autonomous robots in public spaces increases—whether navigating campus walkways or neighborhood sidewalks—understanding how to carefully study these robots becomes critical. While HRI research has conducted field studies in public spaces, these are often limited to controlled experiments with prototype robots or structured observational methods, such as the Wizard of Oz technique. However, the autonomous mobile robots we encounter today, particularly delivery robots, operate beyond the control of researchers, navigating dynamic routes and unpredictable environments. To address this challenge, a more deliberate approach is required. Drawing inspiration from public realm ethnography in urban studies, geography, and sociology, this paper proposes the Walk-Along with Robots (WawR) methodology. We outline the key features of this method, the steps we applied in our study, the unique insights it offers, and the ways it can be evaluated. We hope this paper stimulates further discussion on research methodologies for studying autonomous robots in public spaces. 
\end{abstract}


\section{Introduction} 

As the HRI field emphasizes naturalistic interaction~\cite{vsabanovic2020we}, efforts have been made to integrate robots into our daily lives. These efforts extend from the home to workplaces and public spaces (e.g., airports~\cite{tonkin2018design}, hospitals~\cite{law2021case}, shopping malls~\cite{kanda2009affective}, museums, hotels~\cite{pan2013listening}) and further into urban pedestrian roads and streetscapes~\cite{dobrosovestnova2024we}. A prime example of robots operating in these environments is the autonomous sidewalk delivery robot. In urban contexts, robots navigate sidewalks and streets not as confined to designated ``places,'' but rather as entities traversing multiple locations and zones~\cite{woo2020urban}. They coexist with a diverse yet unpredictable set of actors and variables, such as obstacles and weather conditions. Additionally, these contexts are entangled with various stakeholders, including robot users, vendors, robots companies, city planners, and transportation laws and regulations.

Commercial autonomous robots are expected to become increasingly common not only in public spaces but also on sidewalks, as regulatory and technological constraints continue to relax~\cite{marks2019robots,garland_2023,blanco_2021}. This shift presents researchers with more opportunities to access, observe, and study these robots. Given the unpredictable variables and events, as well as the diverse actors with various purposes coexisting on urban sidewalks~\cite{rosenthal2020forgotten}, it is timely to discuss research methodologies for studying robots in these uncontrolled environments. For instance, when a robot is moving, which pedestrians should be observed?  Where should researchers position themselves on the sidewalk they are observing? (e.g., should researchers position themselves as actors belonging to that sidewalk, blending in as pedestrians?)  Spatially, what should be considered the scope of the robot’s interactions? (e.g., how close should they be to the robot while observing it? To what extent should researchers follow the robot?). For how long should the robot be observed? (e.g., temporally, where do the boundaries of the robot’s interactions begin and end?)

In this paper, we introduce Walk-along with Robots (WawR) methodology, which serves as a conceptual framework and a practical guide for conducting ethnographic fieldwork with autonomous delivery robots in busy urban districts. The WawR has enabled us to plan and implement our fieldwork, including mapping, participant observation, autoethnography, and intercept interviews. Central to this methodology is \textbf{\textit{the act of walking alongside robots}}, inspired by mobile methods from urban studies, sociology, and geography, where researchers walk or ride with human informants to understand their movement through urban spaces. By applying this to robots, we gain insights into how they traverse city neighborhoods and interact with the local communities. 
Conceptually, unlike conventional methods, WawR positions robots as active agents in the research process, aligning with a “more-than-human” perspective~\cite{correia:2024more}.  This shift echoes Anna Tsing’s work on non-human entities~\cite{tsing2015mushroom}, framing \textbf{\textit{robots as the starting points for analysis}}, and examining their influence on both human activities and the environments they navigate, and vice versa.



We outline three key aspects of the WawR methodology, illustrated through examples from our fieldwork: \textbf{the people} encountered along the robots' paths, \textbf{the places} robots traverse, and \textbf{the times} when robots operate. Each aspect highlights how the method provides a structured guide to what can be observed and analyzed. Finally, we detail the step-by-step process for applying WawR in the field, evaluations, and practical tips. We hope this introduction of WawR encourages further dialogue within the HRI community on methodologies for studying autonomous robots in public spaces.

\section{Related Work}

\subsection{HRI Studies on Autonomous Sidewalk Robots}

Research on sidewalk delivery robots spans various methodologies, including online surveys~\cite{kannan2021external,abrams2021:theoretical,wang2023my}, YouTube video analyses~\cite{nielsen:2023using,shin2024delivering}, and experimental studies~\cite{arntz2023assessment,singh2023behavior,washburn2022exploring}. Online surveys have investigated technology acceptance models for autonomous systems~\cite{abrams2021:theoretical} and explored how external human-machine interfaces convey navigational intent to pedestrians~\cite{kannan2021external,yu2023your}. YouTube video analyses have examined unguided interactions and interaction breakdowns in public spaces~\cite{nielsen:2023using} and assessed public reactions through video comments~\cite{shin2024delivering}. Research focusing on human roles in last-mile delivery scenarios has evaluated acceptance through interviews and surveys~\cite{puig2023human,nielsen:2023using,law2021case}. Experimental and field studies, while limited to controlled environments, have assessed traffic readiness for autonomous robots~\cite{arntz2023assessment}, human compliance~\cite{washburn2022exploring}, and the effectiveness of robots' audio-visual aids as social assistants~\cite{pahuja2024delivery,singh2023behavior}.

While there has been limited in-the-wild research on mobile robots in urban public spaces and sidewalks—unpredictable and busy environments—there have been a few recent field studies in these settings\cite{hoggenmueller2020stop}. For instance, Bu et al~\cite{bu2024field}. conducted a study deploying trash barrel service robots in public plazas in Manhattan and Brooklyn, emphasizing methodologies that facilitate naturalistic human-robot interaction, such as in-the-wild deployment and the creation of datasets~\cite{bu2024ssup} from videos captured from the robots' perspectives.

Unlike researcher-controlled robots—such as developed prototypes~\cite{bu2024field}—or research settings under the researcher’s control~\cite{kanda2009affective}\cite{schneider2022stop}, such as field studies in relatively short pre-designated areas~\cite{babel2022findings}, commercial autonomous robots are increasingly appearing in public spaces, operating outside the control of researchers. There have been several HRI studies focusing on this emerging phenomenon of commercial autonomous robots navigating sidewalks~\cite{pelikan2024encountering,weinberg2023sharing,dobrosovestnova2022little}. These studies primarily employed an ethnographic approach centered on observation. In addition to observations, social media discourse analysis and autoethnography were used to understand public interactions with these robots. For example, research~\cite{dobrosovestnova2022little} examined people's willingness to assist robots stuck in snow, shedding light on the potential for spontaneous assistance as a strategy to overcome robotic challenges in uncontrolled environments. During a public pilot of commercial autonomous robots on city sidewalks, researchers~\cite{weinberg2023sharing} observed robot-pedestrian interactions and conducted intercept interviews with residents. These efforts revealed challenges posed by the robots, including distractions, obstructions, and accessibility issues. Additionally, a study~\cite{pelikan2024encountering} employing an ethnomethodological approach, combining field observations with video recordings, identified the nuanced and varied responses of individuals toward robots in public spaces, as well as the social dynamics of everyday street life. This research emphasized the importance of capturing the breadth of interactions. Focusing on pedestrians' interactions with robots, both subtle and overt, these previous studies have provided insightful findings on “what’s going on” in urban pedestrian streets~\cite{dobrosovestnova2024we}.

\subsection{Mobility Studies in Urban Context}

Outside of the discipline of HRI, there have been several attempts to understand the scene of urban settings and the interaction between its actors in the public realm with different research methods and ideas. One notable discussion is the new “mobilities paradigm” \cite{sheller2006new}\cite{cresswell2010towards} or “mobilities turn” which shifts its focus from sedentarist aspects of places to understand how movement constitutes (and of) societies and relations of actors. This attempt integrates several theoretical approaches, including the stream of ideas on mobilities that scholars tried to emphasize the importance of connection, tempo, and a new precision in city life~\cite{sheller2006new}. Moreover, Sheller and Urry~\cite{sheller2006new} discuss the importance of Science and Technology Studies in developing the new mobilities paradigm, which underscores the outcomes of the connection between non-humans and humans in terms of mobility.    

Urban anthropologists Jaffe and de Koning~\cite{jaffe2022introducing} emphasize the characteristics of urban ethnographic methods, indicating how they can be implemented to grapple with profound levels of dedication in everyday lives. These methods, shaped by the \textit{spatial turn}—which recognizes space as embedded with social context and power dynamics—include mobile methods for studying urban lives~\cite[p.19]{jaffe2022introducing}. Mobile methods specifically involve accompanying participants and observing to understand the intricate mechanisms of urban mobility and its actors~\cite{jaffe2022introducing}.






Go-along method~\cite{kusenbach2003street} is a good example of mobile methods, which focus on walking or moving with participants and interviewing to comprehend the everyday experiences of the participants. Sociologist Kusenbach~\cite{kusenbach2003street} argues that go-along enables ethnographers to grasp participants’ perceptions of their environment and introduces researchers to unanticipated settings~\cite{kusenbach2003street}. Similarly, Vestergaard, Olesen and Helmer~\cite{vestergaard2016act} examine the act of walking within the new “mobilities paradigm”~\cite{sheller2006new}\cite{cresswell2010towards} and how this mundane practice embodies cultural context and social norms by focusing on the Danish pedestrian culture. Reed and Ellis~\cite{reed2019movement} extend this, demonstrating how walking as an ethnographic method reveals the banal practices of participants' everyday lives, and provides insights into their interactions with various places and mobilities~\cite{reed2019movement}.


Mobilities scholar Jones~\cite{jones2021public} explored the challenges of conducting ethnographic research in urban public spaces, where traditional methods such as participant observation may not be suitable. He highlighted the value of mixed-methods approaches, which include focused observations of social phenomena, intercept interviews, and fieldnotes that produce reflexive data. In a related discussion,  Wilson~\cite{wilson2020campervan} adopts a non-human-centered perspective by incorporating Actor-Network Theory (ANT) into her study of campervan mobilities in the UK. ANT, which emphasizes the role of non-human actors in everyday life and network formation, enabled Wilson to reconceptualize movement not as “a linear trajectory from A to B”~\cite[p.134]{wilson2020campervan} but as a complex network of actors and influences. These perspectives collectively inspire the WawR, which integrates mixed-methods and non-human-centered approaches to study robots’ movements and interactions in urban spaces.



\section{Overview of Field Sites}

Our study was conducted in two field sites, referred to as Field A and Field B, both located in Seoul, South Korea. The municipal governments in both areas were supporting delivery robot projects as part of broader initiatives to strengthen the country’s robotic competitiveness and develop smart cities (e.g., the ``robot streets'' initiative). In these fields, different robot companies, in partnership with the government, were operating their own delivery robots. Since July 2023, following a two-year exemption from road traffic regulations granted by the Ministry of Land, Infrastructure, and Transport to improve domestic autonomous driving technology and facilitate the commercialization of delivery robots, these robots have been operating in several regions, including Seoul and other large cities~\cite{baek_2023}. Field A and Field B were selected because, after more than a year of continuous operation, they represented areas where interactions between robots and their surroundings had become more stable and established, rather than being influenced by novelty effects.

\begin{figure}
    \centering
    \includegraphics[width=6.7cm]{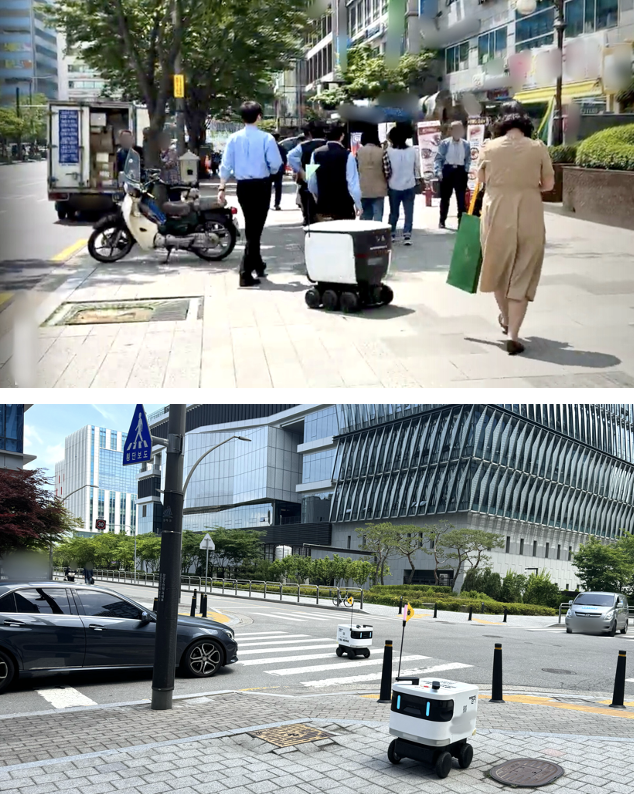}
    \caption{Images captured during fieldwork (Top: a robot after a crowd of people passed by; Bottom: a vehicle stopped in front of a robot at an intersection without traffic lights)}
    \label{fig:fieldwork}
\end{figure}
Field A (service area: 4705 m\textsuperscript{2}) is situated in central Seoul, a bustling area known for its high density of commuters and visitors (approximately 650,000 daily). This area is characterized by a fast-paced flow of people, many of whom are glued to their phones as they move quickly from one location to another. Field A and its participating stores were primarily located in a large indoor shopping mall. Here, indoor robots would first deliver food from the store to a designated spot. Then, an employee hired by the robot company would transfer the food to an outdoor robot on the ground floor, which handled the final leg of the delivery. Field A consisted of five participating stores and six delivery pick-up points, all located across from the robots' departure points, requiring the robots to navigate a crossroads with traffic signals during their deliveries. Typically, 2-5 robots operated during working hours, accompanied by a robot company employee monitoring from a distance. These robots stood about 720 mm tall, with a storage capacity of 25.6 liters. Most customers were regulars.


Field B (service area: 184,737.4 m\textsuperscript{2}) is located on the outskirts of Seoul, a newly developed district where advanced R\&D companies have recently begun to settle, along with an increasing number of new residential complexes. Field B was primarily populated by residents and employees from nearby companies, shops, and restaurants. Unlike the faster pace of Field A, people here moved more leisurely, often in small groups, likely co-workers. Residents were also seen walking dogs or chatting outdoors, contributing to the relaxed atmosphere. The roads in Field B, though occasionally uneven and obstructed by ourdoor banners, and motorbikes cutting through traffic, were navigable. Field B had 10 participating stores and 31 pick-up points, all located on the ground floor. Store staff would come outside to load food into the delivery robots, which waited outside the storefronts. The robots in this field measured approximately 730 x 550 x 730 mm. During our observation, 6-8 robots were docked at their stations, with 2-6 robots operating simultaneously at any given time.


We conducted a week of fieldwork in each site, typically from 8:30AM to 5PM, aligning with the robots’ operating hours (9AM-4PM) on weekdays. Both sites consisted of pedestrian paths free from street vendors or other significant obstructions. Following Hubbard and Lyon’s~\cite{hubbard2018introduction} discussion of walking the streets as a method of embodying urban environments, we paid particular attention to the movements of people, cars, and robots to fully sense the dynamics of each field.




\section{The Walk-Along with Robots (WawR)}

The Walk-along with Robots (WawR) methodology, proposed as a way to study public autonomous robots, literally involves the practice of walking alongside robots to conduct research. 

The WawR offers HRI scholars a structured entry into ethnographic study in urban public spaces, moving beyond ad-hoc observation methods~\cite{dobrosovestnova2022little}\cite{weinberg2023sharing} typically limited to what falls within a researcher’s immediate view. As Jones~\cite{jones2021public} observed, conducting ethnographic research in fluid urban environments poses challenges due to the dynamic nature of these spaces and frequent movement of subjects. The WawR addresses this by integrating observation with mapping, familiarization, auto-ethnography, and interviews. Unlike methods that prioritize shadowing (e.g.,~\cite{pelikan2024encountering}), which engage with observed individuals only briefly and from afar, the WawR emphasizes “engaging with their world” via intensive observation and in-situ interviews during robots’ operational hours. This \textit{continuity} reduces the temporal disconnect often encountered in public space ethnography, allowing observations and interviews to be contextually linked. The WawR also distinctly differs in its ethnographic focus on interconnected networks of human and non-human actors within the field, contrasting with prior HRI studies that primarily focus on human actions and social order~\cite{pelikan2024encountering}. Their human-centered approach limits analysis to short interactions and overlooks robots as independent entities.


A further distinguishing feature of WawR is its foundation in a more-than-human perspective, where non-human actors such as robots are treated as active participants shaping the field. Here, delivery robots are seen as ``interlocutors'' whose paths and interactions with others provide a comprehensive perspective on the environment. 




Drawing inspiration from anthropologist Anna Tsing's work~\cite{tsing2015mushroom}, where she followed matsutake mushrooms—non-human entities—through woodland sources in Japan, Finland, and China, WawR similarly expands our ontological perspectives. Tsing’s work~\cite{tsing2015mushroom}, which revealed relationships between humans, fungi, and trees, offered insights into the transnational social and economic processes that these relationships shape. Crucially, she demonstrated how moving away from human-centered viewpoints deepens our understanding of unnoticed contexts. Similarly, the WawR positions \textbf{robots as the starting and central point of analysis}, broadening the scope to include the various actors, networks, and impacts connected to these interactions. By shifting away from a human-centered view, the WawR enables us to capture more diverse forms of human-robot interaction. This approach places equal importance on how robots affect people, and vice versa. This balance allows for a richer exploration of humans-robot interactions. 

In urban studies, sociology, and geography, mobile methods have become established research methodologies, which involve “walking with” or “riding with” human informants, following and observing their movements through urban spaces. Just as such practices of ‘walking and moving with’ human participants help researchers grasp the logics of movement and the tacit knowledge embedded in the urban landscape~\cite{kusenbach2003street}\cite{jiron2011becoming}, walking alongside robots similarly allows researchers for an exploration of the spaces robots inhabit—neighborhoods within the city and local communities—and the meanings embedded in their movement through these spaces. This includes examining what it means for robots to share sidewalks and public spaces with community members, how the urban landscape influences the robots’ routes, speeds, and modes of operation, and, conversely, how the robot’s presence and movement shape the existing landscape.

In the following sections, we introduce three core aspects of the WawR: \textbf{the people} encountered while following the robot’s path, \textbf{the places} robots traverse, and \textbf{the times} during which robots operate. Each aspect \textit{collectively} contributes to the strengths of this method, offering insights into what can be observed and analyzed.  


\subsection{Encounters with People and the Rhythms Observed While Walking with Robots}


Assume you arrive at your field site to study autonomous public robots operating in the streets. Who will you observe and interview? Will you focus on individuals who directly interact with the robots, or those who encounter them incidentally~\cite{rosenthal2020forgotten}? What specific behaviors will you observe, and what topics will you cover in your interviews? These are essential questions to consider in such a field study.

The WawR presents a way to study the interactions that arise when people and robots share physical space in the field, including \textbf{different forms of human-robot encounters}, \textbf{the norms and flows of walking}, and \textbf{practices related to managing personal space}. It provides a framework for addressing these questions, guiding researchers on what to observe and who to interview, ensuring that the study captures meaningful insights into the interactions between humans and robots in public spaces.

The concept of walking along with a robot is premised on matching the robot’s pace, whether walking closely or from a distance. This allows us to observe \textbf{various subtle interactions}, such as near-collisions, moments of hesitation before encountering a robot, sidestepping at the last moment to avoid contact, and brushing past a robot’s side while walking by. By implementing this method, for example\footnote{The findings throughout this paper are partial, as the focus is on introducing the methodology rather than providing a full analysis.}, we found that humans were the ones who consistently had to give way to robots in our field sites; rarely did the robot initiate a move to avoid people. Instead, the robots prompted people to clear the path, often using auditory cues: either a car horn sound or a polite female voice saying, ``\textit{Hello, I’m delivering. Let me pass by.}'' This consistent pattern revealed that delivery robots occupy a position on the street where they are passively accommodated by people. In essence, the robots assume a role where they are continually granted the `courtesy' of passage, an asymmetrical interaction that reveals how robots command space in human environments.


Secondly, by walking alongside the robot, researchers can observe \textbf{walking norms of the field}, including walking modes and the speed/rhythms of pedestrians. These norms shape the interactions that occur between robots and people. For example, in Field A, a highly crowded region in Seoul, most people walked quickly and maintained a forward-facing posture, often while talking on the phone, keeping pace with those in front of them. As a result, very few pedestrians looked down at the ground, meaning that the delivery robot—roughly knee-high to an adult male—frequently found itself in near-collisions or last-minute avoidance maneuvers as it navigated through this fast-paced walking environment. Almost no one stopped or took time to show interest in the robot. In contrast, in Field B, a district with nearby residential areas and several office buildings that have moved in over the past 5-10 years, the walking mode was more relaxed. People were often seen taking their time, and it was not uncommon for individuals, whether office workers on lunch breaks or local residents, to stop and observe the delivery robot as it passed by.

Thirdly, we can observe \textbf{the distances people maintain from robots }(personal space) and how these spaces evolve due to the operation of delivery robots. For example, in our field sites, we found that while people maintained distinct personal spaces with one another in Fields A and B, these personal spaces were often absent when it came to robots. People tended to touch robots freely or fiddle with their antennas. In the Field B, we even observed pedestrians casually opening the food compartment lids of delivery robots parked outside restaurants while waiting for food orders. This behavior highlighted the absence of social norms regarding how people should interact with robots~\cite{thomasen2020robots}—indicating that touching robots without permission was not seen as problematic.

Interestingly, people seemed more concerned with maintaining personal space with other humans than with robots, often feeling more distance from other people. For instance, we observed a delivery robot operator monitoring the robot’s journey from a distance but retreating into a nearby shop to avoid being seen when the customer arrived to pick up their order. Restaurant employees or owners, responsible for placing food into the robots’ compartments upon their arrival, expressed feeling more comfortable interacting with the robots than with app-based food delivery drivers. One employee noted that dealing with people all day in the restaurant left them feeling fatigued at the prospect of interacting with even more people, such as delivery drivers. Additionally, a restaurant owner shared that seeing delivery drivers waiting to pick up orders made them feel uneasy, knowing that these drivers were under pressure to deliver orders quickly and often worked against the clock.

The participants for the interviews were \textbf{those encountered while walking alongside robots}, following the direction and path set by the robots. These individuals, who can be seen as \textit{involuntary contributors} to interactions in the field site, became a part of study. Specifically, these people represent more than just passersby along the path from point A to point B; they embody a deeper significance tied to the route. As highlighted by the ``mobility turn'' in urban studies~\cite{sheller2006new}, understanding what takes place along these paths requires examining the social context of their movement. This approach underscores the importance of considering not just physical locations but also the social dynamics that unfold in public spaces shaped by autonomous robots. For example, in our field sites, we met a diverse range of individuals who interact with the robots in various capacities. These included office workers from nearby companies, local residents, and people working along the routes that robots traverse. Among these were leaflet distributors, elderly people collecting recyclables, sanitation workers, food delivery couriers, construction workers at nearby sites, and parking attendants. We also spoke to employees of the robots company responsible for managing, monitoring, and maintaining the robots, as well as restaurant owners and staff partnered with the delivery robot service. These individuals are more than just passersby or bystanders passing through the robots’ paths. Thus, the depth of our interviews extended beyond simply concluding with their first impressions or general thoughts and feelings about the robots. Instead, we were able to ask more nuanced questions about \textbf{how these delivery robots exist within their daily lives and what impacts, if any, they have on their work and routines}. 

Urban studies scholars have pointed out a temporal disconnect between those who are observed and those who are interviewed in public realm ethnography. This disconnect is often inevitable due to the heterogeneous and transient nature of public space users. We propose that the WawR approach has the potential to bridge this gap, naturally linking observed individuals with interview participants. For example, while walking with a robot, we encountered a woman handing out flyers near a crowded crossroad. As the robot approached the spot where she was standing, she gestured at the robot and said, \textit{“Move along quickly.”}  During a long traffic signal, we were able to engage in a conversation with her, discussing how the presence of the robot impacted her daily activities. This seamless transition from observation to interview illustrates how the WawR \textbf{can help overcome the temporal disconnect common in public realm ethnography.}

\subsection{Exploring Paths and Spaces Led by Robots: Robots as Actors, Guides, and Experiencing Their (W)heels}


One of the key aspects of our fieldwork using the WawR is that the research field is determined not by human researchers but by the robots themselves. Instead of defining the study area, we followed the paths set by the robots, asking the question: What does it mean to traverse a space through the eyes of a robot? The neighborhoods we encountered became our research field, guided by the robots' movements rather than human intention. We were introduced to various places based on where the robots led us. This shift was made possible by adopting a more-than-human perspective, acknowledging robots as active participants---actors and guides of the neighborhood---in urban environments. By ``walking in their wheels,'' we sought to better understand the robot's experience in these settings.


Integrating ethnographic methods focused on non-human actors, our findings diverged from human-centered research focusing on humans as key interlocutors. Studies like Dobrosovestnova et al.~\cite{dobrosovestnova2022little} and Pelikan et al.~\cite{pelikan2024encountering} emphasize the human experience, focusing on pedestrians' interactions with robots that appeared in their daily lives, and the relationship between the surroundings and humans~\cite{jones2008exploring}. In contrast, the WawR foregrounds \textbf{how robots experience and navigate these neighborhoods}, while acknowledging human-robot interactions in these spaces. One striking observation was that the robots often took routes that differed from what human researchers might choose—sometimes opting for shortcuts that are not present in map/GPS applications. Additionally, our attention was frequently drawn to objects and obstacles at the robots' eye level (approximately 29 inches), such as pavements and intersections. Through this perspective, we identified seemingly minor but impactful road features, like tactile paving and drainage grates, which notably affected the robots' mobility. This more-than-human view allowed us to reframe our understanding of the urban infrastructure and how it influences robotic navigation.

In our research with WawR, \textbf{the robots' movements delineated the boundaries of our fieldwork} within the neighborhood, allowing us to adapt to the space by maintaining a consistent presence. We became attuned to the locations, paths, and routes segmented by the robots' activity range. Reed and Ellis~\cite{reed2019movement} highlight how mobile methods often yield insights from unexpected journeys beyond anticipated boundaries. In contrast, our fieldwork was contained within the robots’ predictable activity ranges. Unlike mobile methods that follow human participants into unforeseen locations, we remained within the robots' paths, walking ahead or behind the robots to gain varied perspectives—such as noticing previously unseen curbs, new obstacles, or different pedestrian interactions.

Jaffe and de Koning~\cite{jaffe2022introducing} argued that ethnographic methods help researchers to understand the surroundings ``in their full complexity''~\cite[p.5]{jaffe2022introducing}. While walking with the robots, which led us to the robots' designated places, we could dive deeper into \textbf{the complex and political entanglement} between the robots and actors in the neighborhoods, which consist of ``both mundane and highly political practices and narratives''~\cite[p.36]{jaffe2022introducing} as they argued. We talked to people who were a part of the robots' movement, and we could get accustomed to the new neighborhoods and learn some intricate relationships that people had with robots and robots' companies. For instance, 
while walking alongside robots in one neighborhood, we observed they frequented a particular café more than other stores. Investigating this pattern, we spoke with the café owner, who revealed ties with the robot company predating the robots’ introduction. The café had partnered with the company, distributing employee coupons that fostered regular visits. As a participant in the delivery robot service, the café encouraged neighboring stores to join, further strengthening its ties with the company. This finding helped us unravel part of the intricate, reciprocal relationships within the neighborhood, where early alliances with the robot company often shaped interconnected dynamics among participating stores.

\subsection{Existing in the Robot's Everyday Lives: Fieldwork Shaped by Robot Time}

The WawR takes a departure from traditional researcher-driven observation by\textbf{ aligning fieldwork with the robots' operational hours.} Instead of the researcher determining observation periods, WawR tracks the robots’ entire work cycle, from station departure, food pick-up, and to delivery, encompassing the full spectrum of their daily operations. This methodological shift mirrors a perspective that emphasizes robot's role in the environment. Previous studies have often relied on fragmented observations, focusing on short periods (e.g., 1-2 hours) or specific interactions within the robot’s broader activity, which were often dictated by the researcher’s availability. In contrast, the WawR emphasizes \textit{continuous} observation, following \textbf{the robot’s full day of operations} over consecutive days. This extended observation allows for insights into the routine nature of the robots' operations, aligning with urban studies’ emphasis on everyday life.

What can be revealed by observing the entire operational time of a robot? This method allows us to understand \textbf{the robot's role (integration) within its neighborhood}—how it fits into the rhythms of the area over time. For example, by observing the full cycle, we can track changes in delivery patterns based on the time of day or day of the week. Details like which restaurants the robot visits more frequently at specific times, what types of food are most commonly delivered, which customers tend to order repeatedly, and when there are surges or lulls in orders become clearer. We can also see \textbf{how these patterns shift over time, which in turn impacts the paths the robot takes and the people it encounters on those paths.}

As we implemented the WawR in the field, we began to interpret \textbf{the temporal context} of the neighborhood through the schedule of the delivery robots, and how specific times affect interactions with the robot. The people using the streets the robot traverses change depending on the time of day, as do the street’s atmosphere and activity levels. For example, early in the morning, street cleaners quietly pass through, while just before lunchtime, delivery riders dominate the streets. During lunchtime, office workers crowd the area, chatting with their coworkers about robots or trying to interact with robots by waving their hands or even touching the robots, followed later by a lull as trucks deliver supplies to restaurants. By mid-afternoon, children returning from school, parents, residents, and visitors converge in the same space. Children with their small backpacks on their back would wave their hands or guardians would suggest taking pictures of the robots. 




Such temporal shifts also alter how robots are perceived. As office workers head out for lunch, they often view robots as a nuisance—``\textit{It's just in the way on the sidewalk}''—and often ignored it. While later, workers returning from lunch tend to find the robot endearing or amusing, with some even patting the robot as they pass by, saying, ``\textit{Oh, it’s here again!}'' These varied reactions highlight how human-robot interactions are shaped by the time and context of the encounter.
Because the robot’s movement patterns change over time, so do the nature of interactions. For instance, during lunch hour, the robot frequently stopped and honked as it navigated crowded sidewalks, whereas outside peak hours, it moved smoothly. These varying conditions elicited a range of reactions from observers. On a hot afternoon, construction workers, watching the robot’s slow progress, remarked, “\textit{At that speed, the food will spoil.}” Office workers nearby, frustrated with the delay, muttered, “\textit{How can it make deliveries like that?}” Yet, in quieter times, such as early mornings or mid-afternoon, a local shopkeeper observing the robot gliding smoothly along the street commented, “\textit{It’s doing well, moving along nicely.}”

Also, these various interactions between the robots and different actors helped us to de-mystify the existence and perception of robots in the neighborhood by revealing how robots exist in the neighborhood without a notable presence, compared to how the researchers imagined before going into the field. The robots did not draw too much attention from those whose daily lives and work happened on the roads of the neighborhood. Specifically, people who spent much time on the road---whether it is to reach a destination occasionally, deliver packages, sustain their livelihood---were not bothered by the existence of robots. The researchers would not have been able to notice this if the researchers were in the field for only a specific period of time in a day.

Furthermore, our method allowed us to integrate perspectives of the mobilities and spatial turn that were mentioned in the previous section. Acknowledging how movement and space are intertwined with context, observing the everyday lives of the robots for full days helped us \textbf{understand how the robots are a part of the neighborhood and not just a passerby or a visitor.} As Sheller and Urry have argued, places do not stand alone~\cite{sheller2006new}, and to situate the actors and place's relationship, understanding the temporal context of the place and actors is crucial since actors, places and interaction between them change over the course of time.

\section{Implementation of WawR in Field Practices}
In this section, we outline the step-by-step process for applying the WawR in the field, accompanied by specific action items.

Prior to fieldwork, we conducted several preparatory meetings to thoroughly understand the context of each field site. This included desk research on the neighborhoods, their demographic compositions, and the socio-economic factors driving the government-supported delivery robot projects in these districts. We also considered the general public's familiarity with service robots commonly seen in South Korea (e.g., in airports, shopping malls, and restaurants), which provided a more nuanced understanding of the role these robots play in S.Korean society and the specific neighborhoods we studied.


\noindent \textbf{Mapping the Robots' Routes and Neighborhoods:} We began by mapping the operational areas of the robots, identifying potential routes they might follow. Using delivery robot service apps, we gathered data on partner stores and pick-up points. These locations were bookmarked on a widely-used local map app in South Korea, known for providing the most up-to-date information. This allowed us to visualize the routes and assess the surrounding infrastructure, including commercial zones, residential areas, businesses, and public transport connections along the robot paths. In addition, we examined factors that might influence the robots' mobility, such as intersections, traffic lights, outdoor banners, bollards, stairways, and pedestrian density. These observations helped us identify unmarked obstacles and consider how robots navigate or adapt to these challenges.

\noindent \textbf{Familiarizing Ourselves with the Field through Embodied Participation:}
Before beginning fieldwork, we aimed to integrate and familiarize ourselves with the neighborhood where the robots would operate. This preparation involved grappling with the physical environment—navigating operating businesses, sewer covers, sloped pavement, and other obstacles.
Conducting an auto-ethnographic approach~\cite{ellis2011autoethnography}, we embraced our roles as participants, positioning ourselves as potential users of the delivery robots. By placing an order and following the robots from their docking station to the delivery point, we experienced their operations firsthand, embodying roles such as customers, pedestrians, and local actors in the study. Through this process, we noticed that the speed of the robots was slower than that of Starship robots, which primarily operate on U.S. campuses. The delivery robots frequently paused due to various obstacles, such as pedestrians and passing bikes. These observations helped us refine our fieldwork strategies for walking alongside the robots. We learned to maintain or adjust our distance based on the robots' speed, identified whether we could comfortably walk on the opposite side of the road while monitoring the robots, and determined if the pathways were wide enough to accommodate both the robots and our presence. Based on these insights, we were able to develop a more structured plan for the subsequent stages of our fieldwork.

\noindent \textbf{Field Notes and ``Pre-Robot Operation Waiting Areas'':}
During the fieldwork, the research team convened one hour prior to the operational hours of the delivery service to plan the day’s observations. We revisited routes previously taken by the robots and compared them with the paths that pedestrians might typically choose. Throughout the day, we observed and recorded the robots’ route patterns—how they traveled to partner stores and subsequently reached designated delivery points. 
To begin each walk-along with the robots, we positioned ourselves near the robots' docking stations—often at vantage points where we had clear visibility, such as benches or outdoor seating at nearby cafés—waiting for the robots to start their deliveries. Once a robot initiated its delivery, we followed it on foot. After each delivery, the robots returned to the docking station, and we resumed our position, waiting for the next dispatch. This cyclical observation allowed us to study the robots' behavior in detail. When a robot appeared at our designated waiting point, we walked alongside it, taking detailed field notes. These notes included the robot's start and end times, the routes and locations (e.g., affiliated stores, sidewalks, and crosswalks), people it encountered, and brief episodes. To gain a different perspective on how robots were perceived, we occasionally changed our waiting areas while ensuring that the robots’ starting locations remained visible. For instance, while waiting at a dog-friendly café, we noted the interactions between robots and dogs, who often barked at the robots coming closer. At a nearby convenience store bench, we observed robots navigating the path while logistics trucks unloaded goods—an opportunity that allowed us to engage in conversation with the truck drivers, regular observers of the robots. By varying our waiting areas, we encountered a wide range of actors (e.g., building managers, recycling collectors) and witnessed different forms of interaction with the robots.
Once we became accustomed to the robots' operational patterns, we began observing from greater distances, such as across the street. This shift in perspective allowed us to gain a broader view of the operation and to note when and how people recognized the robot, often stepping aside to let it pass. In this phase, we also engaged with individuals who frequently shared the public space with the robots—sanitation workers, delivery riders, logistics workers, real estate agents, shopkeepers, and local residents—to understand their perspectives on the robots' presence in their daily routines (as detailed in IV.A).

\noindent \textbf{Observing from the Robot’s Perspective:}
Placing the robot at the center of observation—following its perspective and movements—is crucial, particularly in urban environments where numerous distractions such as sounds, events, and visual stimuli can easily divert attention. This way, we can capture subtle but important interactions. For instance, by observing through the robot’s perspective, we discovered that the robot honks its horn only when a “person” blocks its path. However, when faced with non-human obstacles, such as outdoor banners, delivery boxes on sidewalks, or even dogs, the robot did not honk or emit a verbal message to request that the path be cleared. This was also true when the robot encountered another delivery robot. These interaction patterns with non-human objects became visible only by focusing on the robot itself—findings that might have been overlooked had we centered our observation solely on human-robot interactions.  Admittingly, observing the world from the robot’s perspective—following its path and attentively analyzing its surroundings—proved more challenging than it seemed. In addition to tracking the robot's movements, we had to remain aware of other factors such as changing traffic lights and pedestrians who might suddenly run into the street without noticing us or the robot. Simultaneously, we needed to stay focused on what the robot was "seeing"—the obstacles and environmental conditions along its route. Observing the features of the urban environment that slow down the robot's movement (e.g., protruding manhole covers) or how it navigates around obstacles in a busy area requires deliberate effort to maintain focus on \textbf{how interactions shift ``from the robot'' to its surroundings}.

\section{Evaluation: Suggestions and Practical Tips}


\textit{\textbf{How Do We Know If the WawR Is Being Effectively Applied?}}
Effectiveness can be assessed by examining the quality of the data collected in the field. Specifically, if the data provides meaningful insights into the “people” encountered by the robot, the “places” the robot traverses, and the “time” during which the robot operates, then the methodology is being implemented well. Additionally, if this collected data adequately addresses the research inquiries posed during the field study, it can be considered evidence that the WawR has been applied successfully.

\textit{\textbf{How Can the WawR Be Used More Effectively in Field Research?}}
To enhance its effectiveness, two key aspects should be considered. First, the WawR works best when multiple researchers collaborate in the field. While ethnographic field research has traditionally been conducted by individual researchers, the dynamic and complex nature of public spaces where numerous actors and situations intersect demands a team approach. By having multiple researchers work together, a more comprehensive set of data can be gathered from different perspectives. For example, in connecting who is observed with who is interviewed, one researcher could conduct interviews with observed participants, while others continue to walk alongside the robot, extending the observation process.
Secondly, a deep understanding of the cultural and social norms in the field is important for interpreting the nuanced layers of observed interactions. In South Korea, for instance, certain behaviors, such as approaching strangers in public, are often shaped by prevailing social norms. Two people, particularly women in their 20s to 40s, approaching passersby may immediately raise suspicion. This reaction is influenced by the common practice of religious missionaries, who typically work in pairs and engage strangers in a soft, polite manner. As a result, people tend to avoid them or respond coldly, based on the assumption that they are missionaries, even before any conversation begins. For instance, when our research team (composed of two women) approached a delivery app rider resting on a bench between gigs to request an interview, he initially mistook us for religious missionaries and brusquely told us not to bother him as he was busy. Only after we clarified that we were not proselytizers did he apologize and agree to participate in the interview, revealing how deeply ingrained these cultural associations are in shaping initial interactions. Similarly, taking notes in a physical notebook while walking in public can trigger unintended reactions where such behavior is often associated with investigative activities. People may assume the researcher is an investigator or journalist, potentially drawing unwanted attention and making pedestrians uncomfortable. We observed this during our fieldwork and realized that using a smartphone to take notes discreetly is a more effective approach. This approach would allow researchers to blend seamlessly into their surroundings, as they appear to be using their phones like any other pedestrian, thereby avoiding unnecessary attention.

\section{Conclusion}


In this paper, we introduced the WawR as a practical approach for studying autonomous robots roaming in urban and public settings.
Over two weeks of fieldwork in two areas of Seoul, we used WawR to observe how robots engaged with various interlocutors—humans, animals, and inanimate objects such as sewer caps and uneven pavements. Noting the importance of the mobilities turn and shifts of perspectives, we integrated qualitative research methods, and we could situate ourselves in the context of robots’ encounters, place, and time. This method revealed diverse interactions that highlighted the importance of studying robots' everyday lives in context, which may not have been noticeable when focusing solely on human actors. Our step-by-step guide for implementing WawR invites other researchers to walk along with autonomous robots outside of laboratory settings for deeper understanding of their roles in dynamic urban public spaces.


\bibliographystyle{IEEEtran}
\balance 

\bibliography{reference}




\end{document}